\documentclass[conference]{IEEEtran}
\IEEEoverridecommandlockouts
\usepackage{cite}
\usepackage{amsmath,amssymb,amsfonts}
\usepackage{algorithmic}
\usepackage{graphicx}
\usepackage{textcomp}
\usepackage{xcolor}
\def\BibTeX{{\rm B\kern-.05em{\sc i\kern-.025em b}\kern-.08em
    T\kern-.1667em\lower.7ex\hbox{E}\kern-.125emX}}
\begin{document}

\title{ML2SC: Deploying Machine Learning Models as Smart Contracts on the Blockchain\\
}

\author{
\IEEEauthorblockN{Zhikai Li\IEEEauthorrefmark{1}\thanks{}, Steve Vott\IEEEauthorrefmark{1}\thanks{*These authors contributed equally to this work.} and Bhaskar Krishnamachari}
\IEEEauthorblockA{\textit{Viterbi School of Engineering, University of Southern California, Los Angeles, USA}\\
\{leol, svott, bkrishna\}@usc.edu}
}

\maketitle

\begin{abstract}
With the growing concern of AI safety, there is a need to trust the computations done by machine learning (ML) models. Blockchain technology, known for recording data and running computations transparently and in a tamper-proof manner, can offer this trust. One significant challenge in deploying ML Classifiers on-chain is that while ML models are typically written in Python using an ML library such as Pytorch, smart contracts deployed on EVM-compatible blockchains are written in Solidity. We introduce Machine Learning to Smart Contract (ML2SC), a PyTorch to Solidity translator that can automatically translate multi-layer perceptron (MLP) models written in Pytorch to Solidity smart contract versions. ML2SC uses a fixed-point math library to approximate floating-point computation. After deploying the generated smart contract, we can train our models off-chain using PyTorch and then further transfer the acquired weights and biases to the smart contract using a function call. Finally, the model inference can also be done with a function call providing the input. We mathematically model the gas costs associated with deploying, updating model parameters, and running inference on these models on-chain, showing that the gas costs increase linearly in various parameters associated with an MLP. We present empirical results matching our modeling. We also evaluate the classification accuracy showing that the outputs obtained by our transparent on-chain implementation are identical to the original off-chain implementation with Pytorch.
\end{abstract}


\vspace{-0.3cm}

\section{Introduction}
The advent of machine learning (ML) models in various domains has raised significant concerns regarding privacy and the verifiability of computational outputs. Blockchain technology, renowned for its capacity to preserve node privacy while ensuring transactional transparency across a distributed network, presents a promising solution to these challenges. Recent scholarly efforts have ventured into federated learning algorithms on blockchain platforms, effectively decentralizing the model training process while safeguarding data privacy. This intersection of blockchain with AI for federated learning and data preservation has been the subject of extensive research. 

% Notable studies in this realm include the development of frameworks \cite{b5} for transmitting model weights across blockchain networks for training purposes, thus maintaining the confidentiality of the underlying data.
% The application of blockchain in model training is widely acknowledged for its contributions to data privacy. However, the utilization of blockchain for model inference provides an added layer of transactional transparency. 

Executing models on-chain ensures the authenticity and transparency of the inference outcomes, affirming that they are generated by the designated models. This aspect is increasingly vital as AI systems grow more complex and sophisticated, where reliable verification methods for model inference are essential to mitigate concerns about potential model hijacking, censorship or manipulation.

Despite these advantages, deploying model inference on blockchain platforms faces significant challenges, primarily due to the computational limitations inherent to such systems. The Ethereum Virtual Machine (EVM), a predominant computation engine in the blockchain domain, lacks support for floating-point operations and standard exponentiation functions. Furthermore, the EVM is constrained by system limitations, including a maximum stack size of 1024 and a cap on stack data members at 256 bits. These constraints render the execution of large-scale, computation-intensive AI models impractical.

Building upon prior works such as the one by Badruddoja et al.~\cite{b3}, our study aims to evaluate the feasibility and efficiency of translating simple ML models to run on blockchain platforms. Specifically, we focus on the multi-layer perceptron (MLP) architecture, chosen for its simplistic structure, wide applicability, and relatively modest computational demands. Our research not only assesses the practicality of on-chain MLPs but for the first time introduces an open-source translation mechanism from PyTorch to Solidity\footnote{Github link: https://github.com/ANRGUSC/ML\_onChain}, facilitating the automatic deployment of MLPs on blockchain. Further, we present a detailed mathematical modeling of gas costs as a function of model architecture parameters and verify it with an empirical evaluation. Finally, we show that with the use of an appropriate floating-point library the performance of the on-chain model during execution is identical to that of an off-chain instance. This study aims to bridge the gap between theoretical machine learning models and their real-world applications on blockchain systems, contributing to the ongoing discourse in the field.

We are offering ML2SC as an L1 solution to provide security, verifiability and transparency of the ML models that are deployed on chain. Our contributions can be summarized as follows:
\begin{itemize}
    \item We introduce a Pytorch to Solidity Machine Learning Perceptron Translator
    \item We provide Experimental Results on Pytorch vs Solidity MLP model accuracy
    \item We provide Experimental Results on Solidity MLP Gas Costs
    \item We introduce Ethereum Gas Cost Equations dependent on various MLP architectures
\end{itemize}

\vspace{-0.1cm}
\section{Related Work}

\subsection{On-Chain Data Privacy}

Two innovative approaches stand out in applying blockchain technology to data privacy. The first, ModelChain \cite{b5}, focuses on privacy-preserved machine learning within cross-institutional healthcare predictive modeling. This method employs healthcare data on blockchain nodes and iteratively trains a model across nodes with the least training accuracy. The second approach, developed by Christian Schaefer and Christine Edman \cite{b4}, introduces a hybrid blockchain architecture tailored for enhancing data privacy and transparency. This framework securely logs the access and processing of personal data, and supports data verification through access to a private blockchain. This research underscores the potential for transparent and secure data handling in enterprise contexts.

\subsection{Inference on Chain}

The development of model inference on blockchain platforms has been tackled only by a handful of papers that are directly related to our work. Harris and Waggoner~\cite{b2} presented a framework allowing for continuous updating and sharing of machine learning models. By hosting models on smart contracts, the framework provides free public access and inference, along with incentive structures for high-quality data contributions. Building on this foundation, a subsequent paper by Harris~\cite{b1} conducted an in-depth analysis of the framework, evaluating various machine learning models and configurations, with a focus on the Self-Assessment incentive mechanism. By examining three different models across varied datasets, the research assessed aspects such as model accuracy, user balances, and transaction costs on the Ethereum blockchain. This study also contributed open-source Solidity implementations of specific models (but not a general Python to Solidity model translator).

Badruddoja et al.~\cite{b3} address the implementation of machine learning algorithms within blockchain smart contracts, focusing particularly on the Naive Bayes algorithm. They addressed the challenge of the absence of floating-point data support in environments like Ethereum by developing a novel method that employs Taylor series expansion for probability estimation in integer arithmetic, reducing gas costs at the expense of reduced accuracy. 
Kadadha et al.~\cite{b6} show a specific machine learning model that is trained off-chain and deployed on-chain to predict worker behavior in a crowdsourcing application, providing a concrete use-case for on-chain model inference. 

While all of the above-mentioned works have tackled the same problem of how to deploy Machine learning models as a blockchain smart contract, they do not (a) provide a standard tool to translate models that are typically written by ML developers in Python using libraries such as Pytorch to Solidity smart contracts, and further, they do not (b) provide a general mathematical modeling of how model architecture-related parameters affect the gas cost associated with deploying and running inference on-chain. Our work aims to fill this gap in the literature. Unlike the prior work~\cite{b3} which developed an approximation based on Taylor series expansion resulting in lower performance, we take a different approach, using a more powerful high-precision (signed 59.18-decimal) fixed-point math library called PRBMath~\cite{b8}, which incurs higher cost but gives identical performance to off-chain models.  

\section{Implementation}
The design of ML2SC involves three major components as shown in figure 1: local machine MLP models, PyTorch-Solidity translator, and on-chain model data loader. This section will offer an overview of each module's structure and functionality. 

\begin{figure}[ht]
\centering
\includegraphics[scale=0.65]{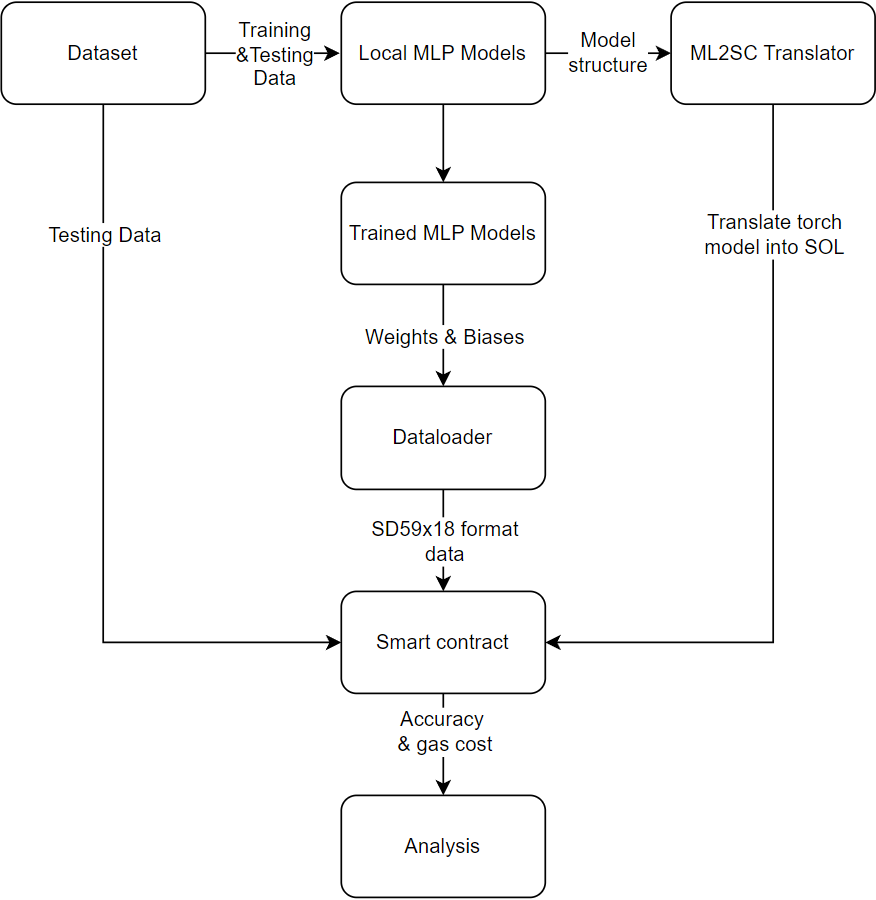}
\caption{Overview of the code design}
\label{fig:experiment_design}
\end{figure}

\subsection{Local MLP Models}\label{AA}
\begin{figure}[h]
\centering
\includegraphics[scale=0.35]{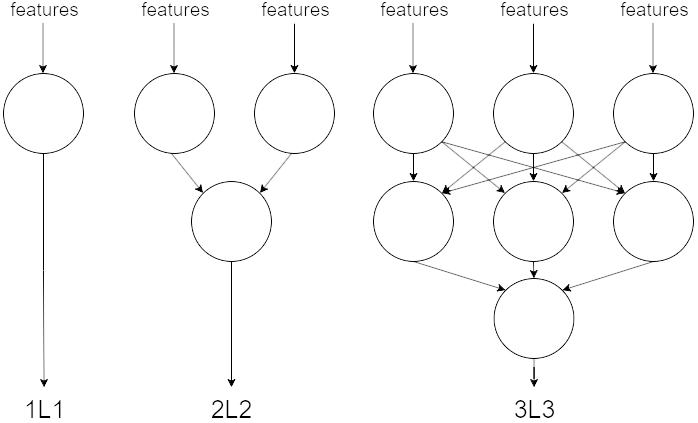}
\caption{Structure of models}
\label{fig:model_graphs}
%\label{fig:weighBiase_upload_gas}
\vspace{-.4cm}
\end{figure}

The system first allows an ML developer to implement any Multi-Layer Perceptron (MLP) and train it off-chain using the Pytorch library. For evaluation purposes, we developed 1, 2 and 3 layer MLP models with different architectural configurations as shown in figure~\ref{fig:model_graphs}. For model $x$L$y$, $x$ is the total number of layers and $y$ is the number of neurons on each layer. However, the system is capable of handling MLP models with any number of layers and any number of neurons in each layer. 

%we implement construct a diverse set of  models to thoroughly evaluate the efficacy and flexibility of our proposed ML2SC system. These models are designed with varying complexities to simulate a range of real-world scenarios and to test the robustness of our translation process. Specifically, for evaluation, 

% \vspace{-0.5cm}

\subsection{ML2SC Translator} 
At the core of ML2SC is a novel PyTorch-to-Solidity translator. This tool is designed to address the challenge of bridging the gap between on-chain neural network evaluation and off-chain source code. Taking into account that most ML developers write models in Python using a library such as Pytorch, our current implementation allows the automated translation of Multilayer Perceptrons (MLPs) of any size written in Pytorch, and outputs the corresponding Solidity code. The generated solidity code has methods for activation functions (sigmoid and ReLu), to set weights and biases, to provide input data, and then to call inference on the whole model.  One challenge for model implementation as a smart contract is that Solidity natively supports only integer arithmetic. Prior work~\cite{b3} therefore tried to approximate calculations with Taylor series expansions, which results in lower accuracy. We take a different approach, using a more powerful high-precision (signed 59.18-decimal) fixed-point math library called PRBMath~\cite{b8}.%cite this

%The PyTorch code for MLPs  a striking resemblance to their translated Solidity counterparts. Each MLP in PyTorch typically comprises the initialization of linear layers with specified dimensional arguments, as well as a simple forward pass function. 
%By introducing the PyTorch-to-Solidity translator, we aimed to facilitate a seamless transition and integration between these two distinct environments, enhancing the efficiency and applicability of neural network models within blockchain frameworks. 

%(added a reference to the cnn paper here, talk about how we used prb to increase precesion)
\vspace{-.2cm}
\subsection{Dataloader}
A JavaScript (JS) data loader is built to parse and upload weights, biases, and test data to the SOL contracts. It has built-in functions that convert floating point numbers to signed 59.18-decimal fixed-point number to work with the  PRBMath library we use. 
%It interacts with the contract through the Truffle Suite test network. 

\vspace{-.4cm}

\section{Experiments}
The experimental framework was structured in three distinct phases: data preprocessing, local model training, and on-chain model execution.

\subsection{Data Preprocessing}
For the purpose of evaluating model performance, we utilized the Heart Attack Analysis and Prediction Dataset\cite{b9}, a binary classification dataset. This dataset comprises 30 numerical features alongside binary target labels. To enhance model performance and ensure data uniformity, normalization techniques were applied to the dataset.

\subsection{Local Training and Inference Export}
In the initial phase of model training, the dataset was partitioned into training and test subsets, with the test set comprising 10\% of the total data. This proportion was strategically chosen to minimize the computational load during the on-chain execution phase. Subsequently, the MLP models of various sizes denoted by wLxN (where w represents the number of layers and x the number of neurons in the input and hidden layers) were trained on the training subset. The accuracy of each model was documented. Following the training process, the weights and biases of these models were extracted and archived into JSON files for further use.

\subsection{On-Chain Classification}
In the final phase, the pre-trained models' weights, biases, and corresponding test datasets were uploaded to their respective smart contracts through the JS data loader. These contracts were then tasked with performing classification tasks using the uploaded parameters. The output of this phase included the classification accuracy and the total gas consumption, providing a comprehensive overview of the model's on-chain performance.

\section{Results}
\subsection{Model Accuracy}

\begin{figure}[ht]
\centering
\vspace{-.4cm}
\includegraphics[scale=0.37]{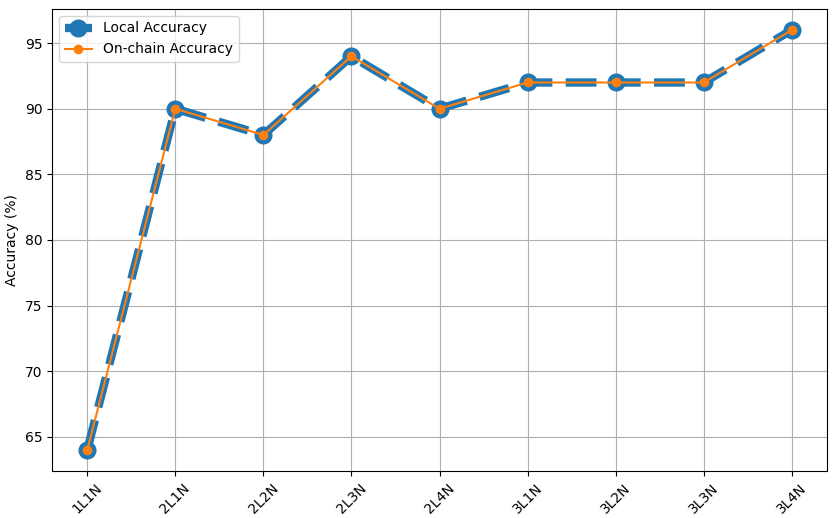}
\vspace{-.2cm}
\caption{Model Accuracy Comparison}
\vspace{-.3cm}
\label{fig:accuracy}
\end{figure}

Pytorch uses IEEE 754 single-precision floating point numbers, making it capable of handling weights with large exponents. The weights in our models range from about $-2.73$ to $3.12$. Our evaluation in figure~\ref{fig:accuracy} shows that on-chain models and their off-chain counterparts have identical performance when tested on the same test set that contains 50 data points. It shows that PRBMath's 58.19 decimal fixed-point is more than capable of representing neuron weights and performing calculations on them that match what is obtained with Pytorch. However, it is worth noting that PRBMath uses significantly more gas than Solidity's built-in arithmetic operations, as shown in table~\ref{tab:PRBMathvSolidity}. We model and evaluate the gas costs incurred next.

\begin{table}[htbp]
\vspace{-.4cm}
\caption{PRBMath vs Solidity Built-in}
\begin{center}
\begin{tabular}{|l|c|c|c|}
\hline
& add & mul & div \\
\hline
Built-in  & 215& 297 & 274 \\
PRBMath & 382& 656 & 617 \\
\hline
\end{tabular}
\end{center}
\vspace{-.6cm}
\label{tab:PRBMathvSolidity}
\end{table}

\subsection{Deployment Gas Cost Analysis}

\begin{figure}[h]
\centering
\vspace{-.4cm}
\includegraphics[scale=0.45]{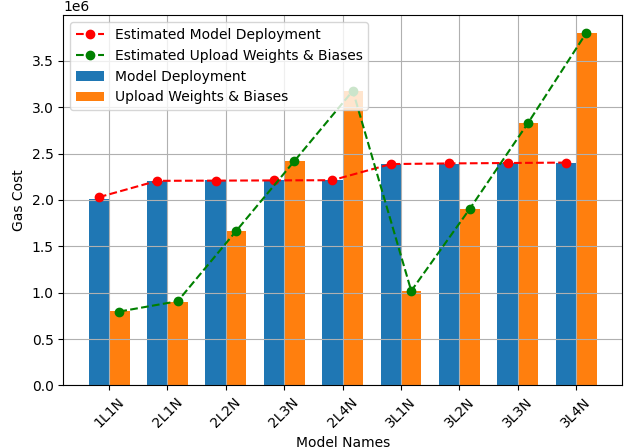}
\vspace{-.2cm}
\caption{Deployment Cost}
\label{fig:deployment_cost}
\vspace{-.4cm}
\end{figure}

The deployment gas cost consists of two major components: model deployment and weights \& biases upload.

The model deployment costs increase as the number of neurons and layers increases. By analyzing the gas cost consumption pattern of different models and testing various parts of the contracts, a general formula was derived to estimate the total deployment/classification cost of on chain MLP contracts. Table II breaks down deployment gas costs into basic building components. Since all models generated with ML2SC are based on combinations of 1L1N - a single perceptron with only sigmoid activation function, more complicated model's gas cost may be calculated using the table and the below steps. 

\begin{table}[h]
\centering
\vspace{-.4cm}
\caption{Model Deployment Gas Estimation}
\begin{tabular}{|l|c|c|}
\hline
Operation & Gas Cost & Symbol\\
\hline
Deployment overhead & 2030000 & $O_D$\\
Each neuron & 2273 & $N_D$ \\
Weights Array new layer & 29320 & $W_D$ \\
Classification new layer & 90000 & $C_D$ \\
Biases new layer & 24987 & $B_D$\\
Set weights/new layer& 32000 & $S_D$\\
\hline
\end{tabular}
\vspace{-.3cm}
\end{table}

The cost of adding a neuron is fixed regardless of the model structure. The gas cost of adding a layer may be roughly broken down into: allocating new weights arrays, updating the constructor, updating set weights function, and updating the classify function. When calculating the deployment cost of ML2SC generated MLP model with $w$ layers, $x$ neurons at each hidden layer, $y$ total number of edges including input layer, and $z$ total number of neurons, the gas cost equals:

\vspace{-.3cm}
\begin{equation}
O_D + (w - 1)\cdot(W_D+C_D+B_D+S_D)+ (x - 1) \cdot N_D\label{eq1}
\end{equation}

The weights and biases uploading cost have a slightly different pattern versus model deployment. Huge jumps in gas costs are observed when adding neurons to existing models instead of adding a new layer. This phenomenon is caused by the multiplying number of weights generated for neurons at the input layer, which accounts for the majority of the upload cost. We estimate these costs empirically using statistical regression.

\begin{table}[h]
\centering
\vspace{-.4cm}
\caption{Weights and Biases Upload Estimation}
\begin{tabular}{|l|c|c|}
\hline
Operation & Gas Cost & Symbol\\
\hline
Weights Overhead & 33164 & $O_W$\\
Each layer& 29963 & $L$\\
Each weight & 22501 & $W$\\
Each biases & 58800 & $B$ \\
\hline
\end{tabular}
\vspace{-.3cm}
\end{table}

The weights and biases upload cost equals:

\begin{equation}
O_W + L \cdot w + W \cdot y + B \cdot z \label{eq2}
\end{equation}

\subsection{Inference Gas Cost Analysis}

Inference cost can be broken down into data uploading cost and classification cost. Data uploading cost is consistent across different models since the same test data set is used. Classification cost generally follows the trend of weights \& biases uploading cost.  

\begin{table}[h]
\centering
\caption{Classification Cost Estimation}
\begin{tabular}{|l|c|c|}
\hline
Operation & Gas Cost & Symbol\\
\hline
Overhead & 3800106 & $O_C$\\
Relu & 22808 & $R$\\
Sigmoid & 28033 & $S$\\
Each edge & 106514 & $E$ \\
Each new layer & 103247 & $L_C$ \\
\hline
\end{tabular}
\vspace{-.4cm}
\end{table}

The classification cost of a model with $w$ layers, $t$ total edges including input layer, and $i$ edges at the input layer equals: 
\begin{equation}
O_C + R\cdot(t-i) + E\cdot t+ L_C \cdot (w-1)+ S \label{eq3}
\end{equation}

\begin{figure}[h]
\centering
\includegraphics[scale=0.45]{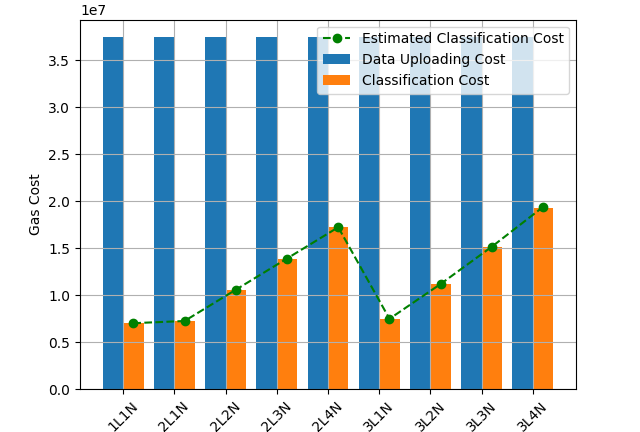}
\vspace{-.2cm}
\caption{Inference Cost}
\label{fig:inference_cost}
\vspace{-.4cm}
\end{figure}

Fig.~\ref{fig:deployment_cost} and Fig.~\ref{fig:inference_cost} show the empirically obtained deployment and inference costs associated with different size models and show that they match the equations provided above.

\section{Conclusion}
Our study successfully demonstrates the integration of Multi-Layer Perceptrons (MLPs) into blockchain frameworks, utilizing a novel PyTorch-to-Solidity translator and a JavaScript data loader. We rigorously tested MLP models of varying complexity, from single-layer to three-layer structures, and found consistent accuracy across all models when deployed on-chain.

A key finding is that there is a consistent pattern between the increase in gas consumption with additional layers and neurons in MLPs on the blockchain; we provide math formulas for the same and show that they match empirical evaluations. This provides a practical guide for deploying machine learning models in Blockchain environments.

While prior works such as \cite{b1, b2, b3, b6} have already shown the feasibility of deploying and running inference on ML models on chain into blockchain systems, our study provides the first automatic translator for any MLP model implemented using Pytorch, to be provided as an open source contribution to the research community. In future work, our translator will be further developed to include more ML model architectures such as CNNs, LSTMs etc.  We also aim to provide a tuning knob for developers to tradeoff accuracy for gas cost. Finally, it is also of interest to investigate alternative L1 and L2 blockchain platforms that offer more cost-effective and accurate blockchain-based machine learning solutions~\cite{b7} moving beyond the current constraints encountered on the Ethereum network.

\section*{Acknowledgement}
This work was supported in part by a grant from AFOSR.


\newpage
\begin{thebibliography}{00}
\bibitem{b1} J. D. Harris, "Analysis of models for decentralized and collaborative AI on blockchain," in Sep 22, 2020, pp. 142-153.
\bibitem{b2} J. D. Harris and B. Waggoner, "Decentralized and collaborative AI on blockchain," in 2019 IEEE International Conference on Blockchain (Blockchain), 2019.
\bibitem{b3} S. Badruddoja et al, "Making smart contracts predict and scale," in 2022 Fourth International Conference on Blockchain Computing and Applications (BCCA), 2022.
\bibitem{b4} C. Schaefer and C. Edman, "Transparent logging with hyperledger fabric," in 2019 IEEE International Conference on Blockchain and Cryptocurrency (ICBC), 2019.
\bibitem{b5} T. Kuo, J. Kim and R. A. Gabriel, "Privacy-preserving model learning on a blockchain network-of-networks," Journal of the American Medical Informatics Association : JAMIA, vol. 27, (3), pp. 343-354, 2020. 
\bibitem{b6} M. Kadadha et al, "On-chain behavior prediction Machine Learning model for blockchain-based crowdsourcing," Future Generation Computer Systems, vol. 136, pp. 170-181, 2022. 
\bibitem{b7} Z. Zheng et al, "Blockchain challenges and opportunities: a survey," International Journal of Web and Grid Services, vol. 14, (4), pp. 352-375, 2018. 
\bibitem{b8}P. R. Berg, “PRBMath,” GitHub. https://github.com/PaulRBerg/prb-math 
\bibitem{b9} R. Rahman, “Heart Attack Analysis \&Prediction Dataset,” kaggle.com. https://www.kaggle.com/datasets/rashikrahmanpritom/heart-attack-analysis-prediction-dataset
\end{thebibliography}
\end{document}